\documentclass[]{bytedance_seed}

\usepackage[toc,page,header]{appendix}
\usepackage{amsfonts}

\usepackage{minitoc}

\usepackage{booktabs}
\usepackage{multirow}
\usepackage{caption}
\usepackage{arydshln}

\title{MMCORE: MultiModal COnnection with Representation Aligned Latent Embeddings}

\author[*]{Zijie Li}
\author[*]{Yichun Shi}
\author[*]{Jingxiang Sun}
\author[]{Ye Wang}
\author[]{Yixuan Huang}
\author[]{Zhiyao Guo}
\author[]{Xiaochen Lian} 
\author[]{Peihao Zhu}
\author[]{Yu Tian} 
\author[\dagger]{Zhonghua Zhai}
\author[*\dagger]{Peng Wang}

\affiliation[]{ByteDance Seed}
\contribution[*]{Equal Contribution}
\contribution[\dagger]{Project Lead}

\abstract{

We present MMCORE, a unified framework designed for multimodal image generation and editing. MMCORE leverages a pre-trained Vision-Language Model (VLM) to predict semantic visual embeddings via learnable query tokens, which subsequently serve as conditioning signals for a diffusion model. This streamlined design effectively transfers the rich understanding and reasoning capabilities of VLMs into the visual generation process. By obviating the need for deep fusion between autoregressive and diffusion models or training from scratch, MMCORE significantly reduces computational overhead while maintaining high-fidelity synthesis.

MMCORE seamlessly integrates text-to-image synthesis with interleaved image generation, demonstrating robust multimodal comprehension in complex scenarios such as spatial reasoning and visual grounding. Comprehensive evaluations indicate that MMCORE consistently outperforms state-of-the-art baselines across a broad spectrum of text-to-image and single/multi-image editing benchmarks.
}

\date{\today}

\begin{document}
\maketitle
\begin{figure}[hb]
\vspace{-3mm}
    \centering
    \begin{minipage}{0.49\linewidth}
        \centering
        \includegraphics[width=\linewidth]{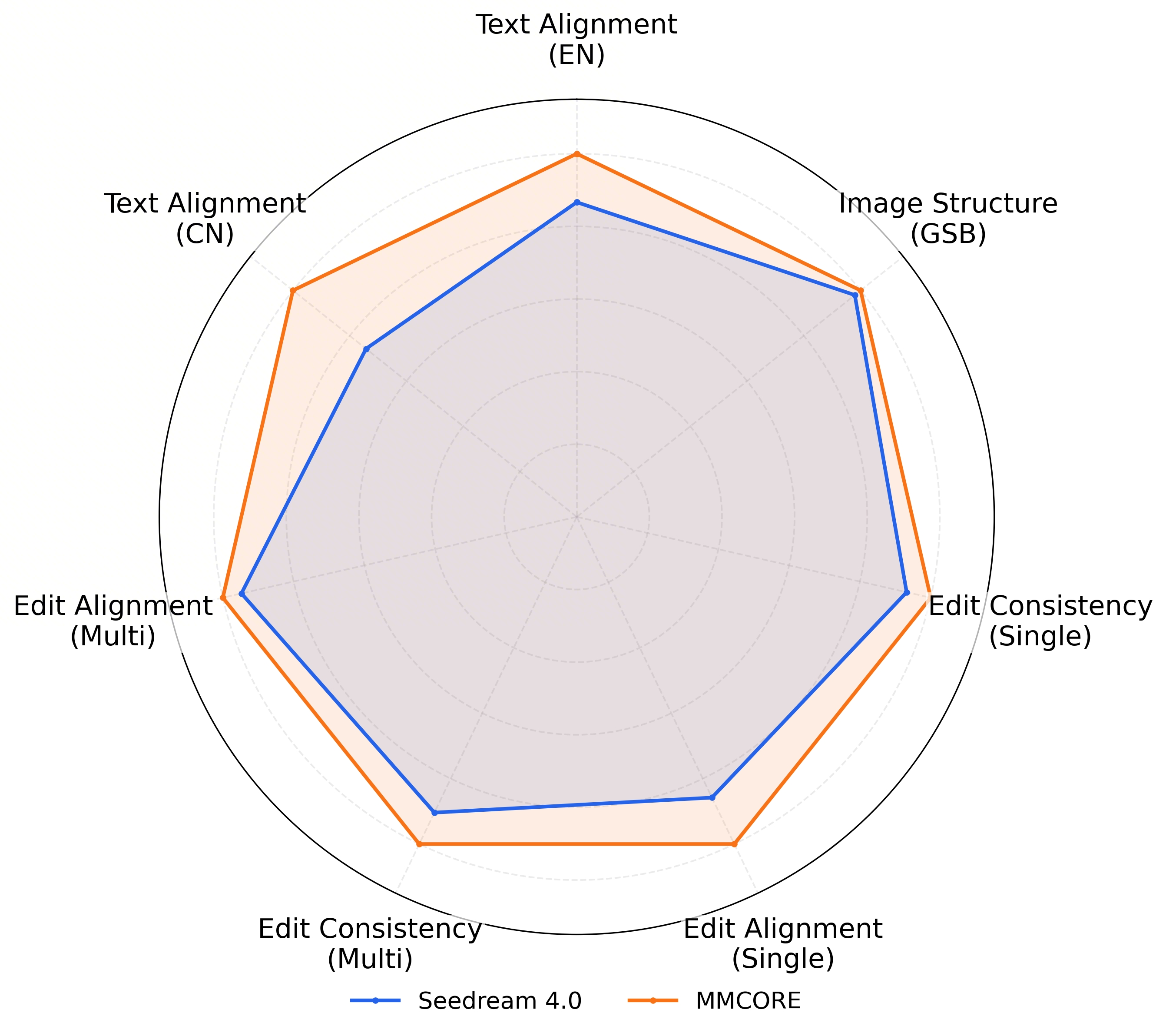}
        \captionof{figure}{Human evaluation over seven metrics.}
        \label{fig:human eval}
    \end{minipage}
    \hfill
    \begin{minipage}{0.50\linewidth}
  \centering
  \centering
  \begin{subfigure}[t]{0.50\linewidth}
    \centering
    \includegraphics[width=\linewidth]{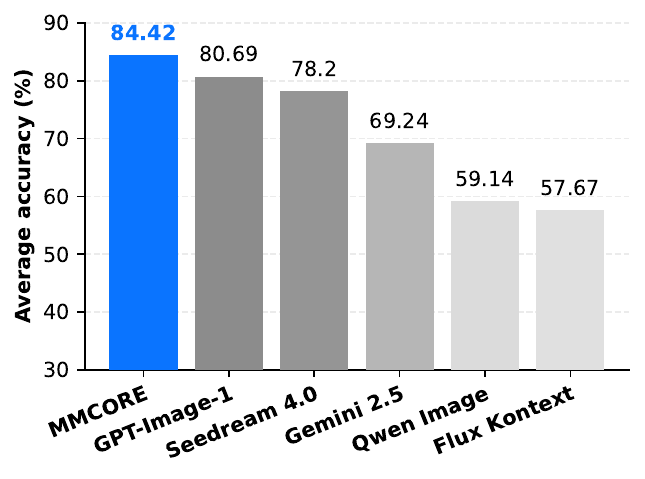}
    \caption{Text-to-image alignment.}
    \label{fig:dreambench_t2i}
  \end{subfigure}\hfill
  \begin{subfigure}[t]{0.49\linewidth}
    \centering
    \includegraphics[width=\linewidth]{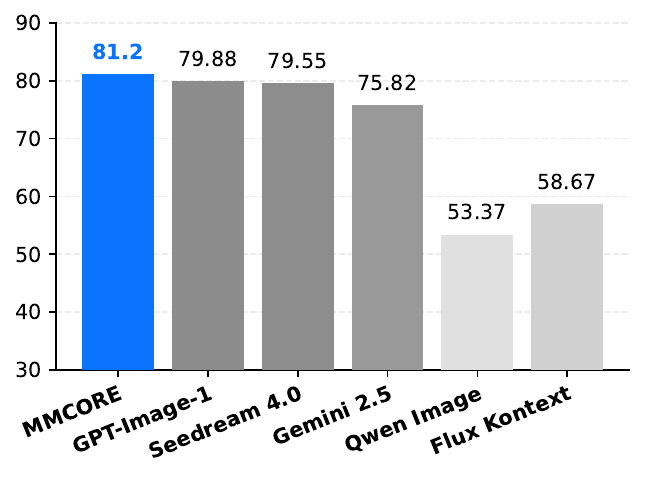}
    \caption{Image-editing alignment.}
    \label{fig:dreambench_edit_align}
  \end{subfigure}
  \begin{subfigure}[t]{0.50\linewidth}
    \centering
    \includegraphics[width=\linewidth]{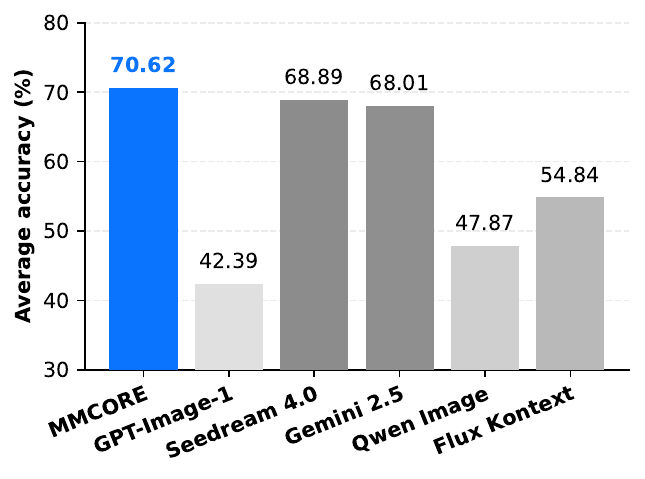}
    \caption{Image-editing consistency.}
    \label{fig:dreambench_edit_cons}
  \end{subfigure}
  \vspace{-2mm}
  \caption{DreamBench AutoEval results of various models.}
  \label{fig:dreambench_three_panels}
    \end{minipage}
\vspace{-10mm}
\end{figure}

\begin{figure}[!htpb]
    \centering
    \vspace{-6mm}
\includegraphics[width=0.95\linewidth]{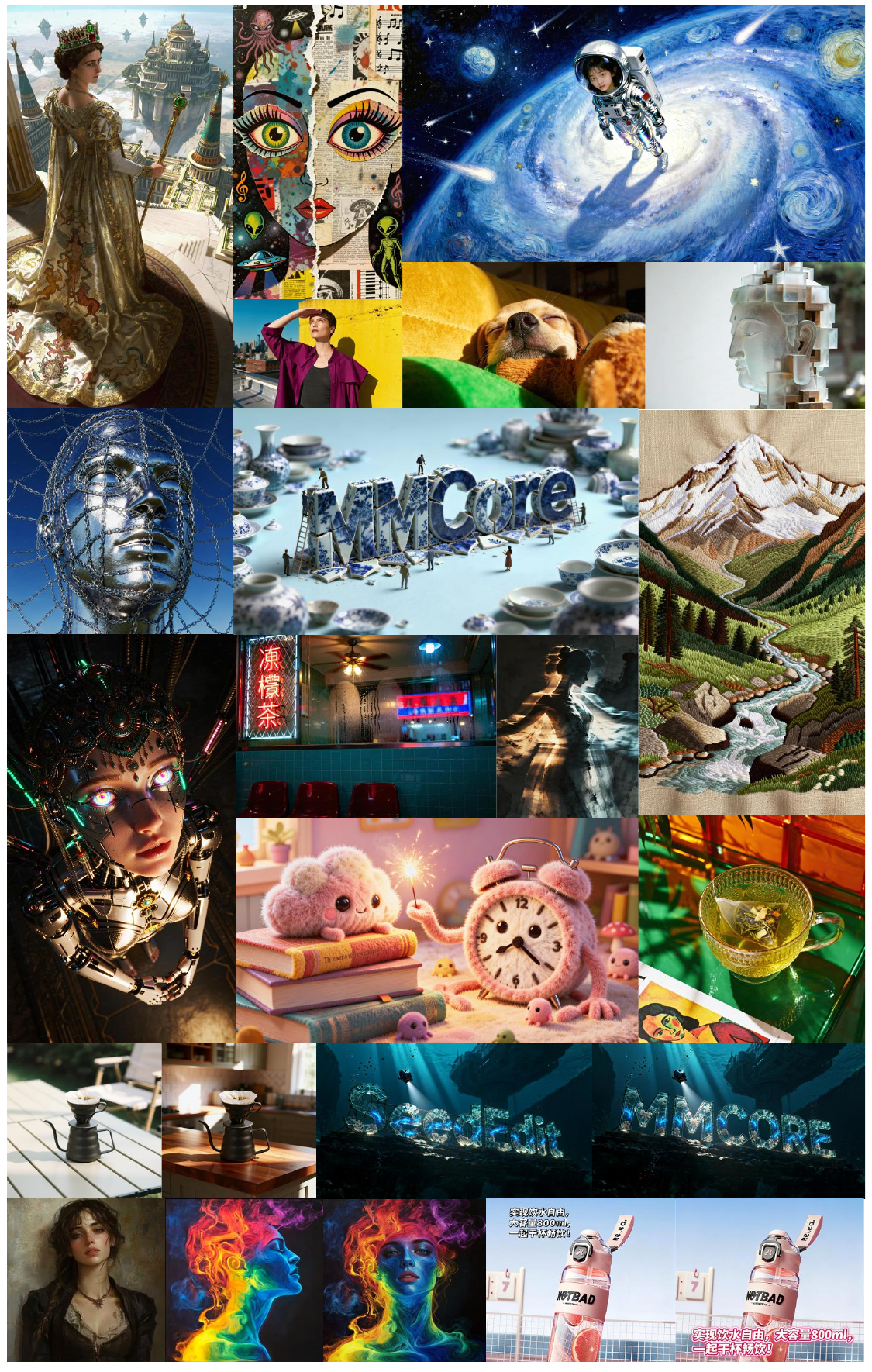}
    \caption{Images generated/edited with the MMCORE model with complex text/image prompts.}
    \label{fig:main teaser}
\end{figure}

\begin{figure}[!htpb]
    \centering
    \includegraphics[width=1.0\linewidth]{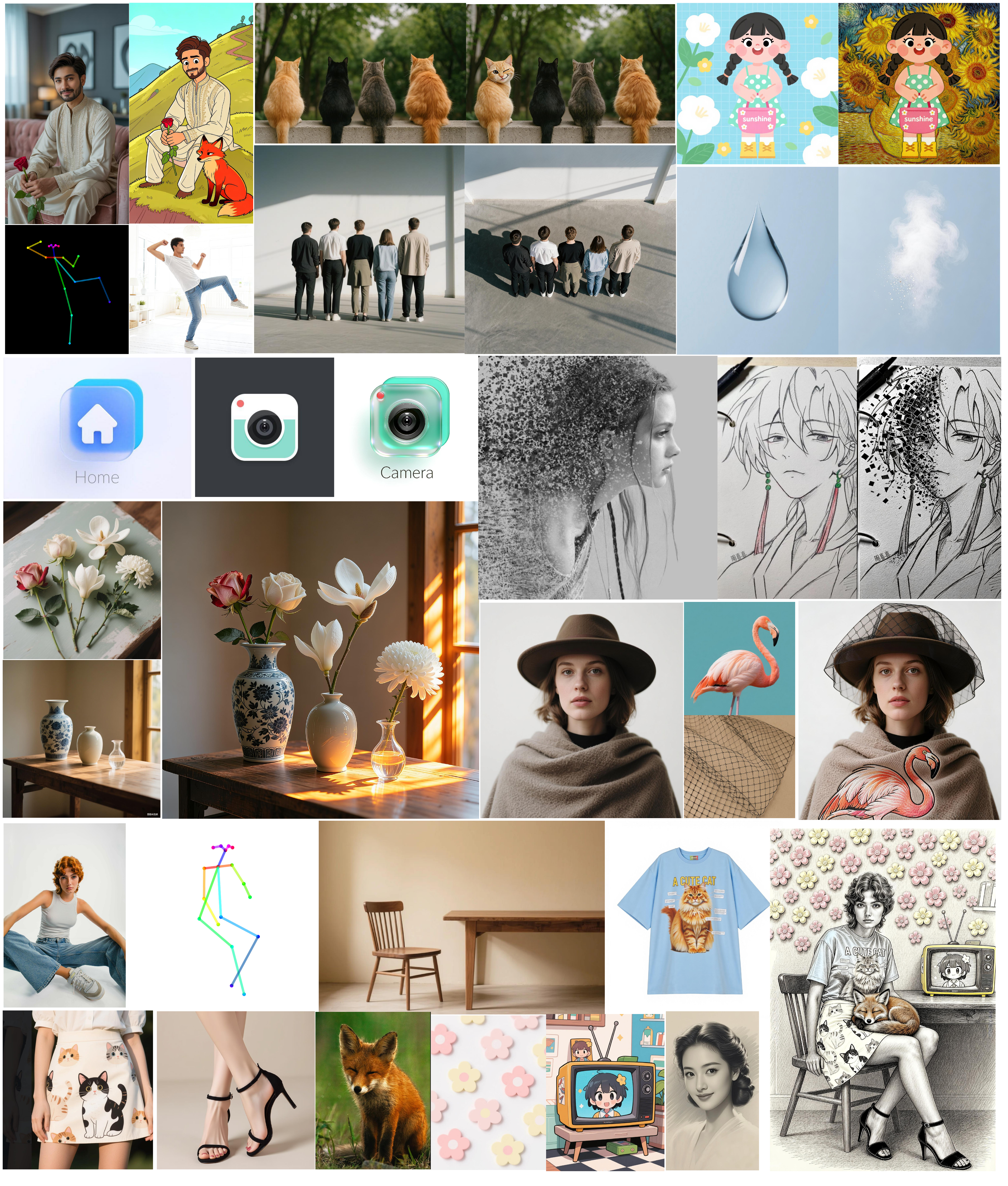}
    \caption{Precise/Reference image editing with MMCORE. Input images are framed at left; output image at right. The model handles multi-image contexts (>10 inputs) with fine-grained localized control.}
    \label{fig:edit_showcase}
\end{figure}

\section{Introduction}


Recently, multimodal generation has emerged as a central problem in machine learning. Two model families dominate current practice: autoregressive (AR) models, which underpin state-of-the-art large language models (LLMs) and vision–language models (VLMs), and diffusion or flow-matching (FM) models, which excel at visual generation tasks such as image and video synthesis. To combine semantic reasoning with high-fidelity visual generation, recent unified models integrate FM mechanisms into AR architectures, as exemplified by Transfusion~\cite{zhou2024transfusion} and BAGEL~\cite{deng2025emerging}.

Directly integrating these paradigms into a single training framework poses significant challenges. In particular, diffusion-based visual modeling degrades training efficiency: unlike text, where a single forward pass supports both understanding and generation, image modeling must alternate between clean features for understanding and noisy features for generation, preventing simultaneous optimization in a single pass~\cite{deng2025emerging}.

To reduce training costs and better align with LLMs, purely autoregressive visual generation models have been explored~\cite{han2025infinity,tian2024visual,li2024autoregressive}. However, their generation quality still lags behind diffusion-based approaches~\cite{seedream2025seedream}, largely due to non-sequential priors and the quantization of visual representations. Other methods, such as MetaQueries~\cite{pan2025transfer}, decouple the two paradigms by using a frozen multimodal LLM to produce conditioning embeddings and by training a diffusion model via a lightweight connector. This strategy preserves modality-specific strengths and efficiently summarizes long-context information, particularly benefiting bidirectional generative architectures such as MMDiT~\cite{esser2024scaling}.

In this work, we propose a multi-stage training framework that achieves a favorable trade-off between generation quality and training efficiency. We first fine-tune an MLLM in an autoregressive manner with causal attention to produce compact latent visual embeddings across diverse multimodal data. The latent visual embeddings are trained to align with semantic representations from state-of-the-art (SoTA) vision–language encoders such as SigLIP~\cite{siglip2}. In the second stage, we train a diffusion model conditioned on text and the latent visual embeddings using flow matching. By simplifying the adaptation of the MLLM, we eliminate the need for explicit connector modules, resulting in a cleaner architecture that is more amenable to post-training procedures such as supervised fine-tuning (SFT) and reinforcement learning (RL).

We evaluate our approach across a range of scaling settings, including loss configurations, batch sizes, and image resolutions. Empirically, incorporating latent visual embeddings enables the diffusion model to leverage richer visual understanding from the VLM beyond text-only conditioning, leading to consistent improvements over state-of-the-art baselines (Fig.~\ref{fig:human eval},~\ref{fig:dreambench_three_panels}). These results demonstrate that high-quality multimodal generation can be achieved by combining autoregressive semantic modeling with diffusion-based visual synthesis, without tightly coupling the two paradigms within a single network.

Finally, our current architecture decouples image understanding and image generation by relying on separate visual encoders. An important next step is to unify these pathways by transforming the latent visual tokens into a sequentialized stream of image-latent tokens suitable for both tasks, enabling both processes to share a single forward pass and further improving model compactness and efficiency.

\begin{figure}[t]
    \includegraphics[width=1.0\linewidth]{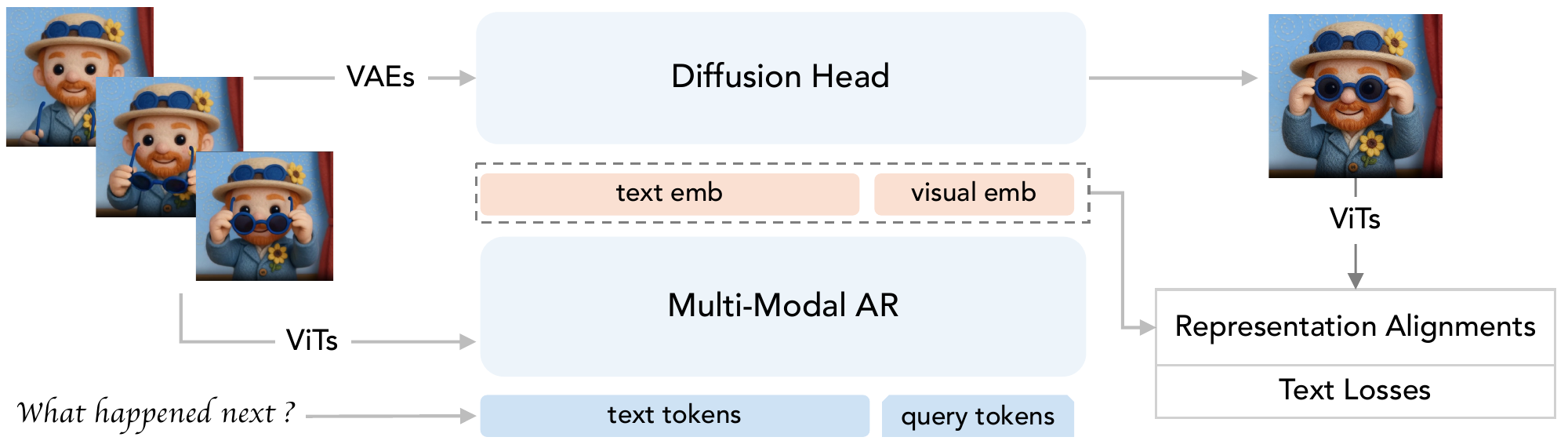}
    \caption{Architecture of MMCORE for low-cost Unified Multi-modal Model. Multimodal information from a VLM is compressed into learned visual latent embeddings, which condition a diffusion-based image generator.}
    \label{fig:architecture}
\end{figure}

\section{Related Work}

\subsection{Unified Multimodal Models}
Unified Multimodal Models (UMMs) have garnered significant attention for their capacity to support both understanding and generation, demonstrating robust generalization across visual understanding, generation, editing, and various downstream tasks. Early UMMs~\citep{team2024chameleon,wang2024emu3,zhou2024transfusion,wu2025janus,chen2025janus,xie2024show} typically utilized a single Transformer backbone to process interleaved image and text tokens. Subsequent approaches introduced separate branches for understanding and generation to ease optimization and improve performance~\citep{shi2024lmfusion,deng2025emerging,wu2025omnigen2,wei2025skywork}. However, these models generally rely on large-scale pre-training over massive image–text corpora, creating significant resource barriers that impede broader research accessibility and exploration.

A parallel line of work, including BLIP3-o~\citep{chen2025blip3}, MetaQueries~\citep{pan2025transfer}, and UniWorld~\citep{lin2025uniworld}, seeks to construct efficient UMMs by coupling frozen multimodal LLMs with trainable diffusion decoders. While this double-fusion structure shares architectural similarities with the modality-specialized mixture-of-experts designs found in LMFusion~\citep{shi2024lmfusion} and BAGEL~\citep{deng2025emerging}, the underlying resource requirements differ fundamentally. The latter models often initialize the generation branch with duplicated LLM weights, necessitating orders of magnitude more training tokens and extensive private data. In contrast, LightFusion~\citep{wang2025lightfusion} trains fused base models—derived from publicly available checkpoints—using strictly public data, achieving competitive performance within a highly efficient and accessible training regime.

\subsection{Cross-Modality Alignment and Representation Transfer}
Cross-modal alignment aims to map heterogeneous visual and textual representations into a shared semantic space that facilitates both understanding and generation. Early dual-encoder frameworks, such as CLIP~\citep{radford2021learning} and ALIGN~\citep{jia2021scaling}, learned modality-invariant embeddings via contrastive objectives. Later multimodal transformers (e.g., BLIP/BLIP-2~\citep{li2022blip,li2023blip2}, PaLI-X~\citep{chen2023pali}) further advanced this by enabling richer vision–language fusion. To reduce computational costs, recent research has explored representation transfer and lightweight alignment strategies for pretrained models~\citep{houlsby2019parameter,hu2022lora,tsimpoukelli2021multimodal}. In generative contexts, diffusion-based models frequently reuse text embeddings to condition image synthesis~\citep{rombach2022high,saharia2022photorealistic,podell2023sdxl}, motivating the direct transfer of multimodal LLM representations to diffusion decoders~\citep{chen2025blip3,pan2025transfer,lin2025uniworld}. Our approach builds upon this foundation, leveraging pretrained multimodal representations and lightweight alignment to enable efficient, high-fidelity image generation without the need for large-scale retraining.
\section{Approach}

In this section, we introduce our unified framework, detailing the model architecture and training strategies designed to align Multimodal Large Language Models (MLLMs) with diffusion-based generators. Given the significant infrastructure gap between current open-source models and proprietary internal state-of-the-art (SoTA) systems, we conduct all experiments using our internal high-performance training pipeline to ensure rigorous evaluation and make our conclusions transferable. 

As illustrated in Fig.~\ref{fig:architecture}, our architecture consists of an MLLM at the base, which infers high-level semantic visual latent embeddings from multimodal and query inputs, and a causal diffusion-based network at the top for pixel-level understanding and generation. During training, target images provide dual supervision: serving as pixel-level targets for diffusion denoising and as semantic-level targets for learning visual tokens.

\subsection{Preliminaries}
MetaQueries~\cite{pan2025transfer} addresses the representational mismatch between the autoregressive latent space of MLLMs and the conditioning interfaces required by diffusion decoders~\cite{ho2020ddpm,rombach2022ldm,peebles2023dit}. The method introduces a set of learnable query tokens $\mathbf{Q}\in\mathbb{R}^{N\times D}$ appended to the multimodal token stream, where $N$ is the query budget and $D$ is the MLLM hidden dimension. Through self-attention~\cite{vaswani2017attention}, these queries distill task-relevant multimodal information into a fixed-size representation, which is then projected into the diffusion conditioning space via a lightweight transformer connector. This mechanism enables the low-interference transfer of MLLM knowledge to high-fidelity image synthesis without requiring end-to-end retraining.

However, relying solely on queried embeddings presents two critical limitations:

\paragraph{(i) Inflexible Context Budget.} A fixed query budget $N$ is suboptimal for handling variable context complexity: a small $N$ fails to capture the nuances of lengthy prompts, while an excessively large $N$ introduces redundancy and computational overhead.

\paragraph{(ii) Weak Alignment.} Supervision derived exclusively from the diffusion objective results in insufficient alignment between the queried representations and the target visual latent space. This leads to reduced training efficiency and heightened sensitivity to data distribution and objective balancing~\cite{rombach2022ldm,peebles2023dit}.

Empirically, we demonstrate that relying solely on query tokens yields significantly weaker performance compared to utilizing the full context. Furthermore, we observe that employing a frozen MLLM backbone results in substantially inferior generation quality relative to its fine-tuned counterparts.

\subsubsection{MLLM for Semantic Visual Prediction}
To overcome the limitations of fixed query budgets and weak supervision, we adopt the query-based token generation strategy from MetaQueries but introduce three critical architectural and training modifications. These changes are designed to bridge the modality gap between the MLLM's text-dominant latent space and the spatially aware representations required for high-fidelity generation:

\paragraph{(i) Joint Fine-tuning of the MLLM Backbone.} Unlike prior approaches that rely on lightweight adapters while keeping the LLM frozen, we fully fine-tune the multimodal backbone jointly on understanding and generation datasets. This enables the model to adapt its weights specifically for visual token synthesis while retaining the core of its pre-trained knowledge. We acknowledge that this aggressive adaptation currently induces a minor regression in general MLLM understanding capabilities. However, we posit that this is a curriculum scheduling issue rather than an intrinsic architectural limitation; we anticipate that introducing generation objectives earlier in the pre-training stage, combined with model scaling, will effectively resolve this interference in future iterations.

\paragraph{(ii) Semantic Visual Alignment via Distillation.} Relying solely on diffusion loss provides a sparse and noisy supervision signal for learning visual tokens. To mitigate this, we introduce explicit intermediate supervision using the latent feature space of a frozen, state-of-the-art vision encoder (e.g., ViT~\cite{dosovitskiy2021vit} or SigLIP~\cite{zhai2023sigmoid}). By regressing the MLLM's generated query tokens toward these semantically dense visual embeddings, we provide a stable, low-variance target. This distillation process acts as a "scaffold," significantly accelerating convergence and ensuring the learned tokens encapsulate robust high-level visual semantics before they are passed to the diffusion head.

\paragraph{(iii) Dual-Pathway Conditioning.} A fixed number of query tokens imposes an information bottleneck, potentially discarding dense textual details required for complex instructions. To address this, we implement a dual-pathway conditioning mechanism: we retain the MLLM's original full-sequence text embeddings alongside the learned visual query tokens. This decomposition allows for a specialized division of labor: the visual query tokens capture global semantics and cross-modal grounding, while the variable-length text embeddings preserve fine-grained lexical details and instruction-following constraints.

Formally, let $\Phi_{\text{vit}}$ be a frozen ViT encoder and $\mathcal{F}_\theta$ be the trainable MLLM. We pool ViT patch features into a $K\times K$ grid to match the query budget ($N=K^2$), yielding target visual features $\mathbf{v}=\Phi_{\text{vit}}(I)$. We optimize the visual tokens via a cosine similarity loss:
\begin{equation}
\mathcal{L}_{\text{vis}} \,=\, \frac{1}{N} \sum_{i=1}^{N}\left(
1 - \frac{\mathcal{F}_\theta(\mathbf{Q}_i)^{\top} \, \mathbf{v}_i}{\|\mathcal{F}_\theta(\mathbf{Q}_i)\| \, \|\mathbf{v}_i\|}
\right).
\end{equation}

The final training objective is:
\begin{equation}
\label{eq:mllm_loss}
\mathcal{L}_{\text{mllm}} =
\lambda_t \mathcal{L}_{\text{llm}}
+ \lambda_a \mathcal{L}_{\text{vis}}.
\end{equation}

Empirically, we find that unfreezing the backbone and explicitly aligning visual tokens consistently improves the quality of image generation and instruction-based editing compared to frozen interfaces, without degrading instruction-following capabilities.

\subsubsection{MLLM Configurations}
\label{subsubsec:mllmconfig}

\paragraph{Two-stage Training.}
As indicated in Eq.~\ref{eq:mllm_loss}, we do not employ a flow-matching loss for the MLLM, allowing it to be trained independently in a two-stage manner. We observe that $\mathcal{L}_{\text{vis}}$ alone is sufficient to adapt the MLLM for semantic feature prediction. Adding flow matching—which would require computationally expensive joint training—yields only marginal improvements while significantly increasing training costs.

\paragraph{Scaling the Query Budget.}
We analyze the impact of the visual token count by varying $N$ from 1 to 128. We find that $N=64$ offers the optimal trade-off between expressivity and efficiency. A smaller $N$ fails to capture sufficient detail for long prompts, while a larger $N$ provides diminishing returns relative to the increased compute. Detailed ablation results are provided in Section~4.

\paragraph{Scaling Batch Size and Sequence Length.}
To maximize training efficiency, we scale our infrastructure from small to large by increasing the number of GPUs, where the batch token length of the MLLM.is expanded. Following the training of a diffusion head on top of the aligned visual tokens, we observe that performance improves consistently with scale, showing no signs of saturation within our maximum compute budget. This aligns with established scaling laws for large-model training~\cite{goyal2017accurate,kaplan2020scaling}.

\paragraph{SFT for Improved Alignment.}
Our decoupled pipeline facilitates staged optimization. Following large-scale pre-training on noisy image–text data, we apply Supervised Fine-Tuning (SFT) using a curated multimodal instruction dataset. A brief SFT phase (e.g., $\sim$2K steps) significantly enhances text faithfulness and controllability, consistent with prior findings in instruction tuning~\cite{ouyang2022training,wei2022finetuned}.

%

\subsection{Modality Mix with Diffusion Head}

\begin{figure}[t]
    \centering
    \includegraphics[width=0.80\linewidth]{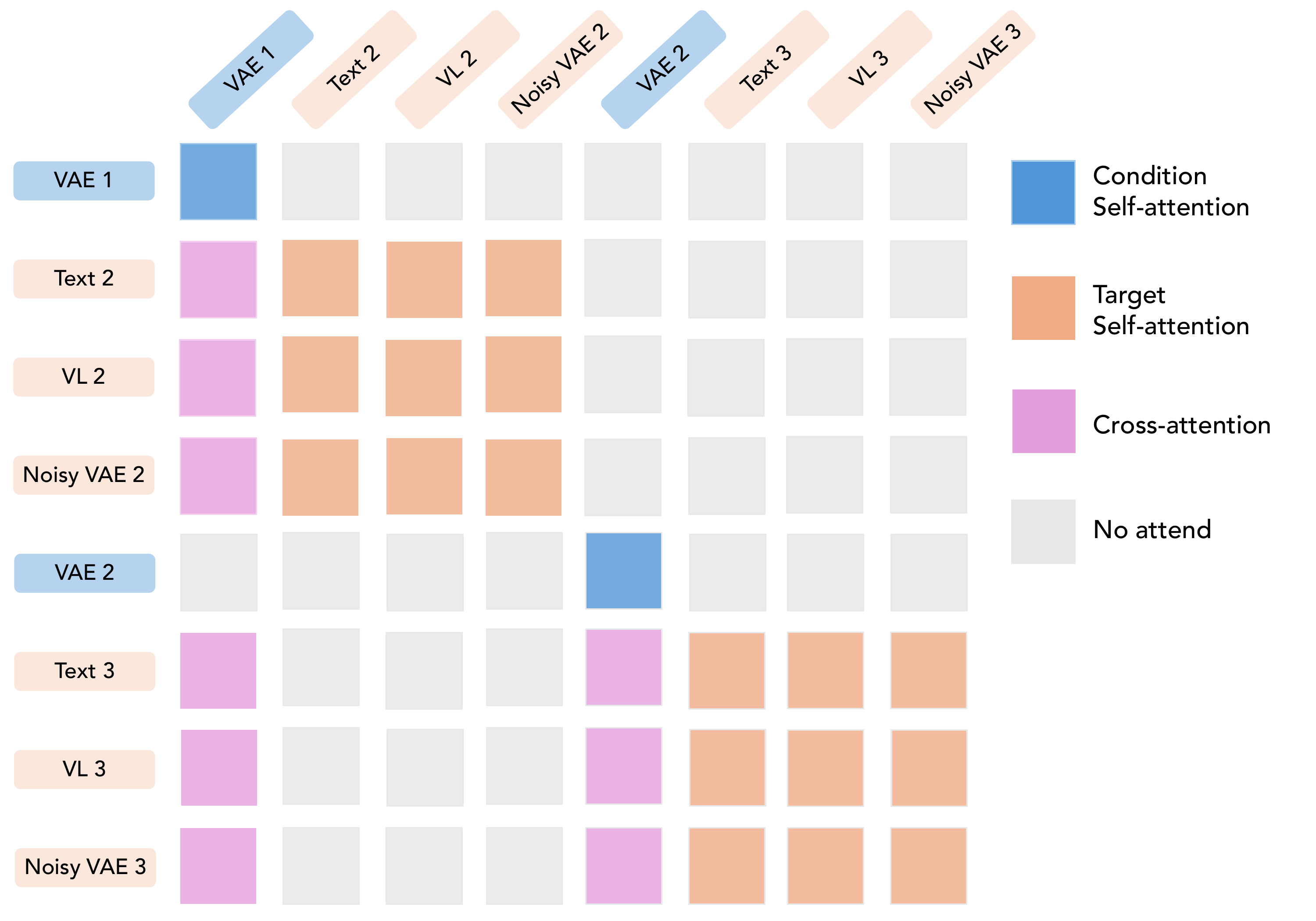}
    \caption{Attention mask for diffusion-head training. Current generation is conditioned on the VAE latents features of all preceding images and the current text/visual latent (\textbf{VL}) embeddings.
    }
    \label{fig:diffusion_forcing_training}
\end{figure}

We leverage a pre-trained Text-to-Image (T2I) Diffusion Transformer (MMDiT) as our generative backbone. Capitalizing on its established text–image alignment capabilities, we extend this model to support multi-reference generation and editing tasks. As demonstrated in prior works~\citep{shi2024seededit,wang2025seededit,seedream2025seedream}, pre-trained backbones can be efficiently adapted to new modalities given high-quality fine-tuning data, significantly reducing the computationally prohibitive cost of training from scratch.

\paragraph{Aligning T2I for Interleaved Generation.}
To handle interleaved multimodal sequences, we re-purpose the model's original self-attention mechanism to support \textit{diffusion-based teacher forcing}, as illustrated in Fig.~\ref{fig:diffusion_forcing_training}. We implement a block-causal attention mask where the generation of the current image frame is conditioned on:
\begin{itemize}
    \item [(1)] The VAE latent representations (via MMDiT features) of all preceding images.
    \item [(2)] The current frame's text and visual latent embeddings.
\end{itemize}
Crucially, we explicitly \textbf{exclude} the semantic visual latent tokens of previous frames from the attention context. Our empirical analysis indicates that attending to historical visual tokens destabilizes the optimization landscape, leading to performance degradation. We attribute this to the complementary nature of the signals: the VAE latents from previous images preserve dense, high-frequency historical details, while the current visual latent tokens provide the necessary high-level semantic guidance for the immediate generation step.

\paragraph{Efficient Training.}
Following these architectural modifications, we proceed to a Continued Pre-Training (CPT) stage using a mixture of T2I and interleaved datasets, while keeping the fine-tuned MLLM frozen. To ensure the diffusion model effectively utilizes both conditioning modalities, we employ an \textbf{Independent Embedding Dropout} strategy, similar to recent approaches in flow matching~\citep{esser2024scaling, flux2024}.

Specifically, we assign independent dropout rates to the text and visual token conditioning pathways. We observe that pre-trained DiT models exhibit a strong inductive bias toward the text conditioning. Thus, we apply a higher dropout rate to text embeddings during the early training stages. This forces the diffusion model to rely more on signals from the MLLM-inferred visual embeddings. This dropout schedule is subsequently annealed to enable robust joint conditioning.

Our streamline integrates multimodal information into the diffusion backbone via meta-query alignment and continued pre-training. In our experiments, we find this strategy is more efficient, achieving performance parity with native unified architectures—such as Transfusion~\citep{zhou2024transfusion} and BAGEL~\citep{deng2025emerging}—while requiring only $\sim$30\% of the computational budget typically associated with training these models from scratch.
 
Finally, keeping the same dropout settings and following the established alignment pipelines~\citep{seedream2.0}, we refine the model via Supervised Fine-Tuning (SFT) on a specialized high-quality dataset, followed by Reinforcement Learning with Human Feedback (RLHF). This final stage aligns the model closer to its theoretical upper bound, yielding a network that significantly outperforms the baselines. 

\paragraph{Discussion: Necessity of VAEs.}
Recent advances, such as RAE~\cite{zheng2025rae}, have demonstrated that discriminative backbones (e.g., ViTs) can potentially support strong image reconstruction. However, current iterations still lag behind VAEs in terms of reconstruction stability and spatial fidelity~\footnote{https://bfl.ai/research/representation-comparison}. Furthermore, the efficacy of these ViT-based approaches has yet to be empirically validated in complex interleaved generation scenarios. We posit that as the reconstruction capabilities of discriminative encoders converge with those of generative autoencoders, the distinction between the two will blur. In such a future regime, the explicit dependency on VAE latents for historical context—and by extension, the necessity of the diffusion head mechanism—may be eliminated, allowing for a fully unified token-based architecture.
\section{Results}
In this section, we illustrate the results to demonstrate the effectiveness of MMCORE. Notice for results comparison, we adopted the same prompt input for ours and Seedream 4.0.

\subsection{Quantitative Benchmarks}


\begin{figure}[ht!]
    \centering
    \includegraphics[width=\linewidth]{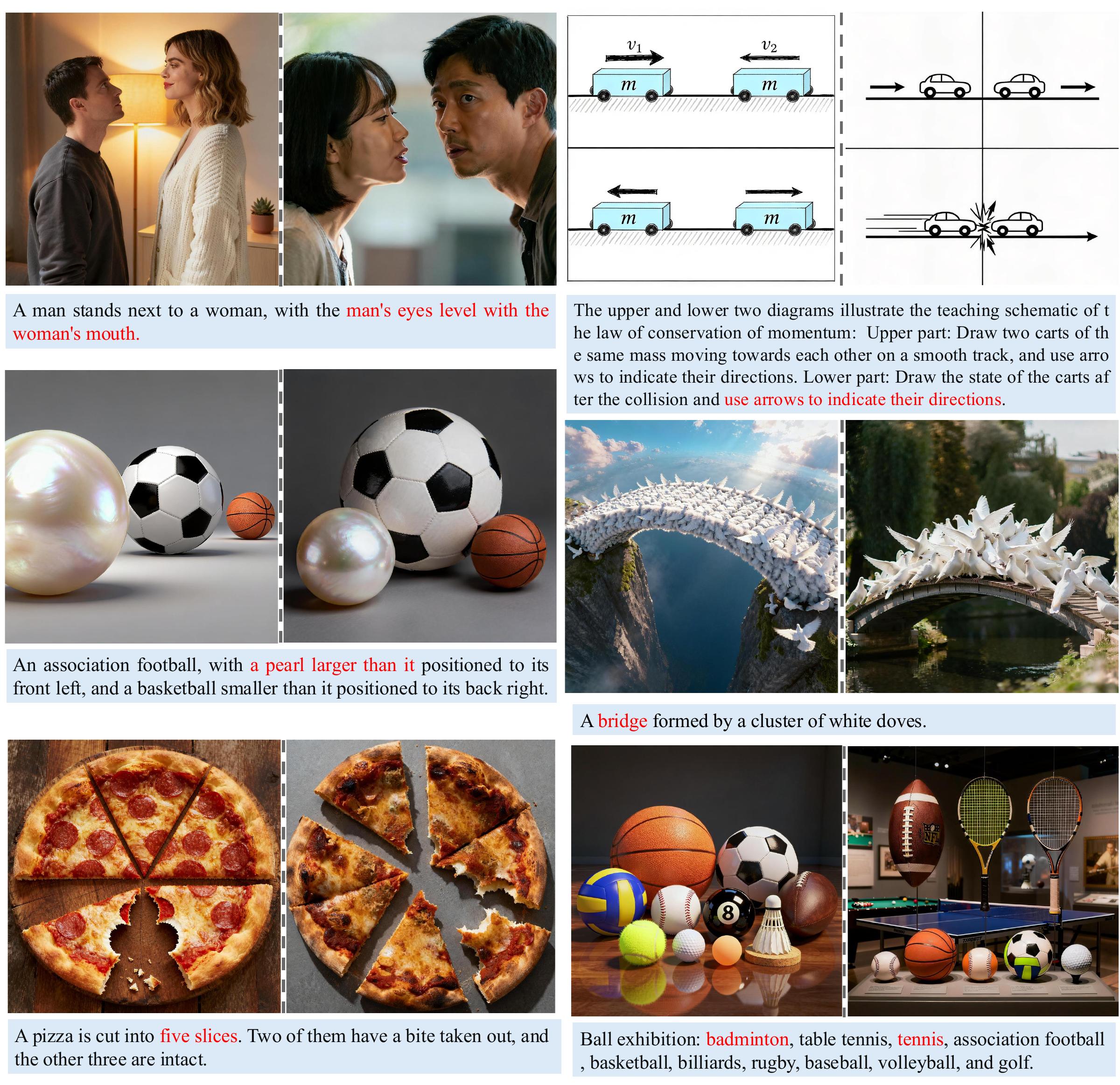}
    \caption{Text-to-image generation comparison against Seedream 4.0. For each pair of images, left: MMCORE; right: Seedream 4.0.}
    \label{fig:t2i_case_vs_seedream4}
\end{figure}

We benchmark MMCORE on an internal suite (DreamBench)~\citep{seedream2025seedream} designed to evaluate both text-to-image generation and image editing via a robust automated pipeline.

\paragraph{Text-to-Image Generation.} As shown in Fig.~\ref{fig:dreambench_t2i}, MMCORE achieves superior prompt-image coherence based on our automatic evaluation metrics~\citep{seedream2025seedream}, significantly outperforming baseline models including Seedream4.0~\cite{seedream2025seedream} and other SoTA methods, thanks to our MLLM's strong text interpolation abilities.

\paragraph{Image Editing.} In editing tasks (Fig.~\ref{fig:dreambench_edit_align},\ref{fig:dreambench_edit_cons}), our model also demonstrates enhanced instruction alignment and consistency, validating the effectiveness of the proposed dual-pathway conditioning, where the ViTs indeed provide an additional semantic visual representation compared to purely diffusion based methods with text/meta-info connection as in prior methods~\cite{wang2025seededit,seedream2025seedream}.

\paragraph{Human Evaluation.} To corroborate automated metrics, we conduct a comprehensive human evaluation (Fig.~\ref{fig:human eval}) across three axes: (i) Prompt–Image Alignment (English/Chinese), (ii) Structural \& Visual Fidelity, and (iii) Editing Consistency (preserving identity and background). MMCORE delivers consistent improvements across all dimensions over our baseline, indicating high reliability in following complex instructions while maintaining visual aesthetics. It also validate the fidelity of our autoeval metrics.

\begin{figure}[t!]
    \centering
    \includegraphics[width=\linewidth]{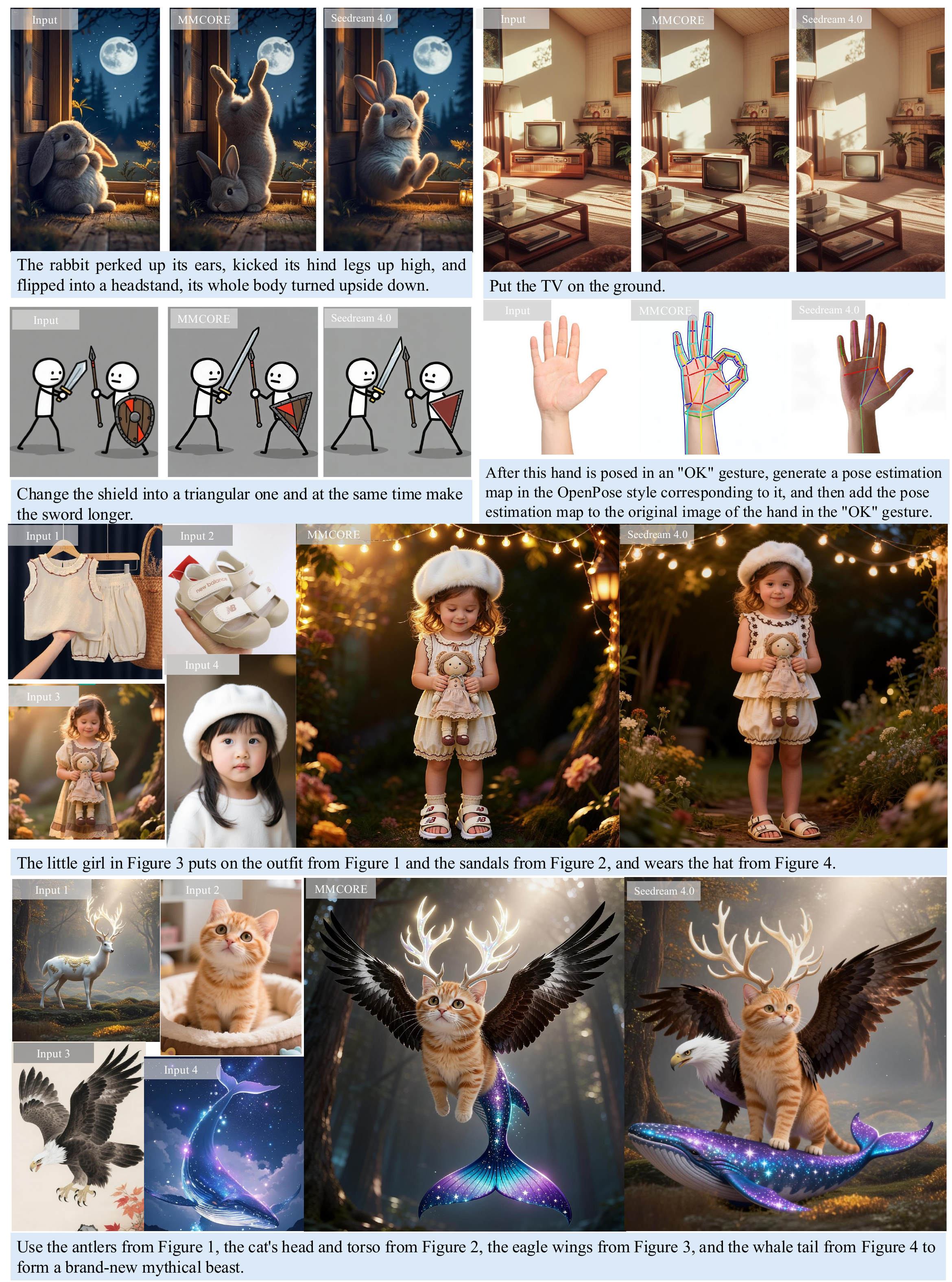}
    \vspace{-6mm}
    \caption{Single- and multi-image editing results comparison against Seedream 4.0. }
    \label{fig:edit_compare}
        \vspace{-10mm}
\end{figure}

\subsection{Qualitative Analysis}
Fig.~\ref{fig:t2i_case_vs_seedream4} highlights MMCORE’s robustness on counterfactual and reasoning-heavy prompts, where baseline models often revert to stereotypical visual priors and lose key textual constraints. Relative to Seedream~4.0, MMCORE demonstrates stronger compositional role binding and attribute grounding: it correctly assigns asymmetric relations (e.g., the man’s eye level aligned with the woman’s mouth), preserves literal interpretations of non-literal phrases (e.g., “a tree skipping rope”), and faithfully instantiates fine-grained, localized details (e.g., a printed orange cat on a shirt with the tail extending along the sleeve). Beyond single-scene reasoning, MMCORE better adheres to dense, multi-part specifications, such as structured grids/collages, indicating improved text fidelity, spatial reasoning, and constraint satisfaction under complex prompts.

Fig.~\ref{fig:edit_showcase} highlights the model's versatility in precise editing. MMCORE generalizes across a wide spectrum of edit types, ranging from single-image semantic modifications (e.g., viewpoint shifts, style transfer) to complex multi-image compositions. Notably, the model scales effectively to long contexts, maintaining fine-grained control even when conditioned on over 10 input images. Fig.~\ref{fig:edit_compare} shows MMCORE’s improved edit controllability and instruction fidelity over Seedream~4.0. It more consistently performs precise spatial and geometric edits (object relocation, position swaps, shape changes) while better preserving non-target content.

\subsection{Ablation Study}
\label{sec:ablation}

To efficiently validate our architectural decisions and training strategies, we employ a lightweight diffusion head for all ablation studies. We evaluate performance using GPT-4o and Doubao-VL\footnote{\url{https://www.volcengine.com/experience/ark?model=doubao-1-5-thinking-vision-pro-250428}} as automated judges to assess semantic alignment and visual quality. Specifically, we design a prompt that instructs these models to rate the alignment between the generated image and the text condition, yielding a normalized score between 0 and 1. Here we reveal a few important conclusions from our study.

\begin{table}[t]
\centering
\caption{\textbf{Ablation on MLLM Architecture \& Tuning Strategy.} Evaluation of connector depth and fine-tuning methods on text-to-image (T2I) generation (Dreambench). Metrics show absolute scores and absolute gains ($\Delta$) compared to the fixed 2-Layer Baseline.}
\label{tab:ablation_connector}
\small
\setlength{\tabcolsep}{10pt}
\begin{tabular}{llcc}
\toprule
\textbf{Tuning Strategy} & \textbf{Configuration} & \textbf{GPT-4o} $\uparrow$ & \textbf{Doubao} $\uparrow$ \\
\midrule
\multicolumn{4}{l}{\textit{1. Connector Depth Analysis (Fixed VLM)}} \\
Baseline (50k) & 2 Layers (Ref) & 0.6791 & 0.7276 \\
& 3 Layers & 0.7490 & 0.7910 \\
& 6 Layers & 0.7843 & 0.8396 \\
\addlinespace[2pt]
& \textit{\color{gray}$\Delta$ vs. Baseline} & \textit{\textbf{\color{gray}+10.5\%}} & \textit{\textbf{\color{gray}+11.2\%}} \\
\midrule
\multicolumn{4}{l}{\textit{2. Full Model Fine-Tuning}} \\
Full Fine-Tuning (50k) & Full Weights & 0.8114 & 0.8417 \\
& Full Weights (Large BS) & 0.8199 & 0.8528 \\
\cmidrule(lr){2-4}
\textbf{+ SFT (2k)} & \textbf{Full Weights (Large BS)} & \textbf{0.8585} & \textbf{0.8915} \\
\addlinespace[2pt]
& \textit{\color{gray}$\Delta$ vs. Baseline} & \textit{\textbf{\color{gray}+25.9\%}} & \textit{\textbf{\color{gray}+16.4\%}} \\
\bottomrule
\end{tabular}
\end{table}


\paragraph{Architecture and Fine-tuning Strategy.}We first established a baseline following the MetaQueries setting~\citep{pan2025transfer}, coupling a frozen VLM with a pre-trained DiT by replacing the original text encoder. Initially, training the query tokens and connectors using only the DiT loss resulted in poor convergence compared to standard text-to-image diffusion models. To address this, we adopted our visual latent alignment loss (Eq.~\ref{eq:mllm_loss}) within a two-stage training framework (Sec.~\ref{subsubsec:mllmconfig}), which significantly accelerated convergence.

With this stable baseline, we examined the capacity required to bridge the modality gap between the MLLM and DiT. As shown in the top section of Table~\ref{tab:ablation_connector}, a shallow 2-layer connector yields suboptimal results (0.6791). Increasing the depth to 6 layers provides a substantial performance boost (+10.5\%), confirming that a high-capacity projector is essential for aligning disparate latent spaces. We further compared parameter-efficient tuning (LoRA~\citep{hu2022lora}) against full model fine-tuning. While LoRA improves upon the baseline, we found that Full Fine-tuning achieves the lowest convergence loss and superior generation quality ($>0.81$), justifying the additional computational cost. Furthermore, scaling the batch size by $5\times$ yielded consistent improvements in text-image alignment, raising the GPT-4o score from 0.8114 to 0.8199.

\paragraph{Effectiveness of Supervised Fine-Tuning (SFT).}Finally, we evaluated the efficiency of a targeted SFT phase (Table~\ref{tab:ablation_connector}, bottom). We observed that extended pre-training (50k steps) exhibits diminishing returns, with performance plateauing around 0.82. In contrast, transitioning to a short, high-quality SFT phase (2k steps) dramatically boosts the alignment score to \textbf{0.8585} (GPT-4o) and \textbf{0.8915} (Doubao). 
 This demonstrates that SFT is indispensable for aligning model outputs with granular human esthetic preferences.


\paragraph{Impact of Visual Latents on Conditional VAEs.}
During diffusion head training, we investigated the potential benefits of augmenting the conditional VAE's DiT encoder features with our learned visual latent embeddings. Leveraging a pre-trained DiT backbone, we found that introducing these additional visual cues resulted in a marked performance degradation. Specifically, our automated evaluation (Doubao for Edit Alignment) revealed a significant drop from 55.2 to 30.62 after 15k fine-tuning steps. Qualitatively, the model exhibited a few failure modes, including severe image artifacts and a tendency to trivially copy reference inputs. These findings indicate that fusing heterogeneous visual features—specifically, dense VAE latents and high-level ViT embeddings—introduces substantial optimization challenges when utilized simultaneously as conditioning signals.

\section{Conclusion and Limitations}
In this work, we present a resource-efficient, streamlined, yet highly effective framework for aligning the long-context understanding and reasoning capabilities of Multimodal Large Language Models (MLLMs) with latent visual representations. By introducing a trainable query mechanism alongside a dual-pathway conditioning strategy, our approach bridges the modality gap without the prohibitive costs of end-to-end retraining. Extensive experiments demonstrate that our method significantly improves image generation fidelity and instruction adherence compared to baseline counterparts, such as Seedream 4.0~\citep{seedream2025seedream}, particularly in complex multi-turn interleaved generation scenarios.

Despite these advancements, we admit there is still a performance gap between ours and Nano-Banana-pro~\cite{google2025nanobanana} / GPT image 1.5~\citep{openai2025gptimage}. We think this might be due to the gap in the MLLM model since adding prompt rewrites from SoTA VLM models, MMCORE performs much better.
Secondly, our current framework exhibits limitations that highlight critical directions for future research:

\paragraph{The Understanding-Generation Trade-off.}
A persistent challenge in current Unified Multimodal Models (UMMs) is the performance degradation observed in pure understanding tasks after generative alignment. Consistent with other SoTA approaches, connecting the MLLM to a generative decoder and fine-tuning for visual synthesis imposes a "tax" on the model's original reasoning capabilities (e.g., VQA, OCR), especially for those highly optimized product models. While our joint training strategy mitigates this catastrophic forgetting, achieving parity with specialized understanding models remains an open challenge. 

\paragraph{Redundancy in Visual Latents.}
Currently, our learned visual tokens function as an \textit{amendment} to text conditioning rather than a comprehensive \textit{supplement} or replacement. This implies a degree of semantic redundancy, where the visual tokens alone are insufficient to drive high-fidelity generation without the auxiliary ViT encoder and diffusion decoder. This limitation underscores the necessity for a next-generation "Omni-Tokenizer"—a unified visual representation capable of supporting both pixel-perfect reconstruction (like a VAE) and high-level semantic reasoning (like a ViT) simultaneously. Developing such a tokenizer to unify understanding and generation at the representation level is the primary focus of our future work.

\bibliographystyle{plainnat}
\bibliography{main}

\clearpage

\beginappendix


\section{Ethical Claims}
The images presented in the paper are from our lisenced ones, and public license-free websites such as Unsplash and Pixabay. In addition, note that the technique proposed in this paper aims to facilitate the user's common tasks that are widely demanded in industry for ethical purposes. It SHOULD NOT be applied to unwanted scenarios such as generating violent and sexual content. It might also inherit the biases and limitations of T2I models. Therefore, we believe that the images or models synthesized using our approach should be carefully examined and presented as synthetic.





\end{document}